\RequirePackage{fix-cm}
\documentclass[smallextended]{svjour3}       
\smartqed  
\usepackage{graphicx}
\journalname{SN social sciences}
\usepackage{authblk}

\usepackage{lineno} 
\usepackage{xcolor, soul}
\sethlcolor{white}
\begin{document}
\title{Hate versus \hl{Politics}:\\Detection of Hate against \hl{Policy makers} in Italian tweets}


\author{Armend Duzha  \and
        Cristiano Casadei \and
        Michael Tosi \and
        Fabio Celli 
}

\institute{Maggioli S.p.A. \at
              Via Bornaccino 101 \\
              47822, Santarcangelo di Romagna \\
              Italy\\
              \email{fabio.celli@maggioli.it}           
              \emph{ORCID 0000-0002-7309-5886} 
}

\date{Received: Dec 23, 2020 / Accepted: Jul 07, 2021}

\maketitle

\begin{abstract}
  \hl{Accurate detection of hate speech against politicians, policy making and political ideas} is crucial to \hl{maintain} democracy and free speech. Unfortunately, the amount of labelled data necessary for training models to detect hate speech \hl{are} limited and domain-dependent. In this paper, we address the issue of classification of hate speech against policy makers from Twitter in Italian, producing the first resource of this type in this language. We collected and annotated 1264 tweets, examined the cases of disagreements between annotators, and performed in-domain and cross-domain hate speech classifications with different features and algorithms. We achieved a performance of ROC AUC 0.83 and analyzed the most predictive attributes, also finding the different language features in the anti-policymakers and anti-immigration domains. Finally, we visualized networks of hashtags to capture the topics used in hateful and normal tweets. 
\end{abstract}


\maketitle

\section*{Acknowledgments}
The research leading to the results presented in this paper has received funding from the PolicyCLOUD project, supported by the European Union’s Horizon 2020 research and innovation programme under Grant Agreement no 870675.

\section{Introduction and Background}
The rise of Natural Language Processing (NLP) tasks \hl{focused on} hate speech \cite{badjatiya2017deep} and the analysis of online debates \cite{celli.al14corea} \hl{have both highlight} bad behaviors in social media, such as offensive language against vulnerable groups (e.g., immigrants, minorities, etc.) \cite{poletto2017hate}, as well as aggressive language against women \cite{saha2018hateminers}. An under-researched - yet important - \hl{area of investigation} is anti-policy hate: the hate speech against \hl{politicians, policy making} and laws at any level (national, regional and local). \hl{While anti-policy hate speech } has been addressed in Arabic \cite{guellil2020detecting}, most European languages \hl{have been under-researched}.\\
In recent years, scientific \hl{research} contributed to the automatic detection of hate speech from text with datasets annotated with hate labels, aggressiveness, offensiveness, and other related dimensions \cite{sanguinetti2018italian}. Scholars \hl{have} presented systems for the detection of hate speech in social media focused on specific targets, such as immigrants \cite{del2017hate}, and language domains, such as racism \cite{kwok2013locate}, misogyny \cite{frenda2019online} or cyberbullying \cite{menini2019system}. Each type of hate speech has its own vocabulary and its own dynamics, \hl{thus} the selection of a specific domain is crucial to obtain clean data and to restrict the scope of experiments and learning tasks.

We have formulated three Research Questions:
\begin{itemize}
    \item RQ1: How different are hate speech domains, such as anti-immigrants and anti-policy? 
    \item RQ2: Is it possible to perform cross-domain training to exploit techniques and models trained in one domain (i.e. anti-immigration) to detect hate speech in another domain (i.e. against policy-makers)?
    \item RQ3: Is it possible to identify and track the topics of public debate involved/not involved in hate speech?
\end{itemize}
\hl{In order to address} RQ1, we performed correlation and classification analysis. The former \hl{was carried out} to measure how different language features are related to hate speech in different domains, the latter to test the performance of classifiers in different domains. To \hl{address} RQ2, we performed cross-domain classification and applied hate speech models trained in an anti-immigration domain to a policy-making domain. \hl{Finally, to address} RQ3, we extracted the hashtags from tweets labelled as hateful and non-hateful, visualized the network of co-occurrences with a Hifan Hu graph \cite{hu2015visualizing}.\\
With this research, we aim to provide actionable insights for evidence-based decision-making \cite{kyriazis2020policycloud}, as online hate is often a predictor of offline crime \cite{williams2020hate}. 
We selected Twitter as the source of data and Italian as the target language for two reasons:\\
1) There are datasets annotated with anti-immigrant hate speech labels in Italian, but no datasets annotated with anti policy making hate speech labels, \\
2) Italy has, at least since the elections in 2018, a large audience that pays attention to hyper-partisan sources on Twitter that are prone to produce and retweet messages of hate against policy making \cite{giglietto2019multi}.\\
This paper contributes to the scientific research in NLP and hate speech detection \hl{in two ways}. \hl{First:} the production of a new corpus, annotated with hate speech labels, in an under-resourced language (Italian). \hl{Second:} the classification of hate speech tweets against policy making, and its comparison to the classification of hate speech against immigrants.\\
The paper is structured as follows: after a literature review (Section \ref{prev}), we collect a stream of tweets in Italian using keywords (i.e., hashtags) related to laws and regulations (Section \ref{data}). \hl{We then} train, test, and evaluate models for hate speech from existing resources, analyze the predictive power of each feature, visualize the results (Section \ref{exp}), and draw conclusions (Section \ref{conc}).

\section{Related Work}\label{prev}
\begin{figure*}[h]
\centering \includegraphics[width=\textwidth]{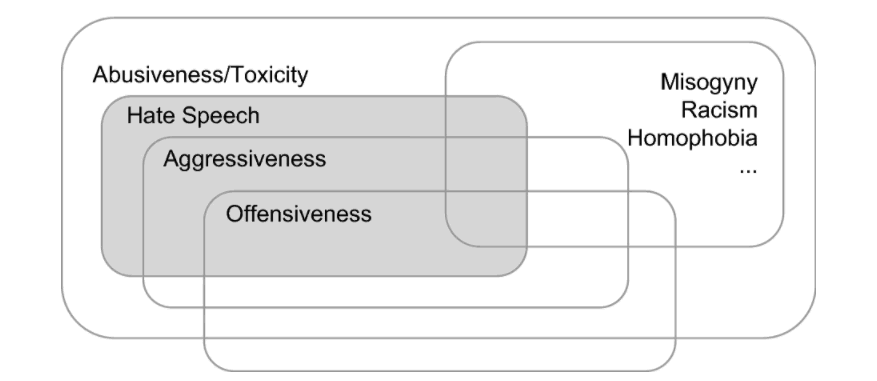}
\caption{\small{Relation between Hate Speech and related concepts. Source \cite{poletto2020resources}.}}\label{hs-rel}
\end{figure*}

Hate speech is defined as any expression \textit{that is abusive, insulting, intimidating, harassing, and/or incites, supports and facilitates violence, hatred, or discrimination. It is directed against people (individuals or groups) on the basis of their race, ethnic origin, religion, gender, age, physical condition, disability, sexual orientation, political conviction, and so forth} \cite{erjavec2012you}. A recent study defined the relationships between hate speech and related concepts (see Figure \ref{hs-rel}), highlighting the fact that involved phenomena make hate speech especially hard to model, with the risk of creating data that is biased and making the models prone to overfitting. \hl{In addition to this, literature also reports cases} of annotators’ insensitivity to differences in dialects and offenses \cite{sap2019risk} that make annotation difficult. For these reasons, one of the \hl{largest} challenges in the field of hate speech is to investigate architectures \hl{that} are explainable, stable and well-performing across different languages and domains \cite{poletto2020resources}.
Another \hl{key} issue is that many recent approaches based on word embeddings \cite{church2017word2vec}, Deep Learning algorithms and BERT Pre-trained transformers \cite{devlin2018bert} \cite{tenney2019bert} \cite{polignano2019alberto}, are vulnerable to undesirable bias in training data, especially in the political domain \cite{wich2020impact}, and suffer from poor interpretability \cite{macavaney2019hate}. In other words, it can be difficult to understand \hl{how} the systems based on Deep Learning techniques make their decisions about hateful/non-hateful messages. Moreover, the decisions taken by systems might be based on biased and unfair models. A method for explaining \hl{the decisions of} transformer models is to look at the attention vectors \cite{clark2019does}. \hl{Yet, studies show that} learned attention weights are frequently uncorrelated with gradient-based measures of feature importance, \hl{thus,} different attention distributions can nonetheless yield \hl{similar} predictions \cite{jain2019attention}. In a context of policy making, the transparency of the decisions and the possibility to interpret the results should be considered a priority.

Despite there \hl{being} many studies about hate speech in Natural Language Processing (NLP) against various targets, such as anti-immigrants, there are few works in the field of hate speech detection against politicians and policy making. Previous approaches to this task exploited transparent Machine Learning (ML) algorithms, such as Gaussian Na\"ive Bayes, Random Forests or Support Vector Machines (SVM), as well as Deep Learning algorithms, such as convolutional neural network (CNN), Multi-Layer Perceptrons (MLP), Recurrent Neural Networks (RNN) with long-short-term memory (LSTM) or bi-directional long-short-term memory (Bi-LSTM) on top of word embeddings extracted from the training set or pre-trained from other resources with transfer learning. These \hl{studies show} that good results can be obtained with Bi-LSTM, MLP and SVM \cite{guellil2020detecting}. 

\hl{Studies} that provided useful datasets in the field of hate speech \hl{include} SemEval 2019 who studies multilingual hate speech against immigrants and women in English and Spanish \cite{basile2019semeval}. In Italian there are two main corpora, both about anti-immigrant hate: the Italian HS corpus \cite{poletto2017hate} and HaSpeeDe-tw2018, the dataset released during the EVALITA campaign in 2018 \cite{haspeede2}. 
The former is a collection of more than 5,700 tweets manually annotated with hate speech, aggressiveness, irony and other forms of potentially harassing communication. The latter, is a dataset (3,000 tweets for training and 1000 for testing) manually annotated with hate speech labels. The results of HaSpeeDe-tw2018, reported in Table \ref{sota}, are the state-of-the-art in hate speech detection in Italian and show that lexical resources, such as polarity and emotion lexica, \hl{are useful to this task} \cite{bosco2018overview}, \cite{fersini2018overview}. 

\begin{table}[h]
\centering 
\begin{tabular}{lll}
\hline
\textbf{algorithm} & \textbf{features} & \textbf{\textit{F1} }  \\
\hline
\hline
baseline & - & 0.403 \\
\hline
SVM, RNN & PolSbjLex, 2 Wemb & 0.799 \\
SVM+RNN+LR meta & ptWemb badWLex & 0.793 \\
SVM & EmoLex, Wemb, social & 0.783 \\
RNN & ptWemb & 0.663 \\
SVM+RNN+RF meta  & ptWemb  & 0.649 \\
\hline
\end{tabular}
\caption{\small{State-of-the-art in hate speech classification. The reported systems use as algorithms: Support Vector Machines (SVM), recurrent Neural Networks (RNN), Linear Regression (LR), ensemble learning (meta) and Random Forests (RF). As features: word-embedding extracted from training set (Wemb) or pre-trained (ptWemb), Polarity/Subjectivity (PolSbjLex), bad words (badWLex), emotion lexica (EmoLex) and social media metadata (social). } \label{sota}}
\end{table}

Most hate speech recognition systems at HaSpeeDe-tw2018 exploit SVM, Recurrent Neural Networks with LSTM or ensemble learning (meta) \cite{bai2018rug}, \cite{michele2018comparing}, \cite{de2018hate}, and word-embeddings as features \cite{santucci2018detecting}, pre-trained or extracted from the training set. Some systems also use cross-platform data (i.e. Facebook and Twitter) and shows that this strategy yields similar results for Twitter \cite{corazza2019cross}. Crucially, the best performing systems make use of lexical resources for polarity, subjectivity and emotions \cite{cimino2018multi}, showing that word embeddings are more effective when combined with lexical resources. \hl{The current state-of-the-art in the HaSpeeDe task from Twitter is 0.808 macro-F1, obtained using transformer-based models} \cite{sanguinetti2020haspeede}.
Regarding the visualization, the heuristic power of network graphs has been known in computational social sciences for at least one decade. For example, network graphs of topics or Twitter hashtags can be used to analyze sentiment polarization of hyper-partisan topics \cite{garimella2017long}. \hl{Another example,} networks of replies annotated with personality types can represent the conversational dynamics of neurotics and emotionally stable users \cite{celli.rossi13}. 

In the next section, we describe how we created the dataset and annotated it with hate speech labels.

\section{Data Collection and Annotation}\label{data}
In order to monitor the reactions of society towards policy making, we retrieved a stream of tweets in Italian from March to May 2020, using snowball sampling. Starting from a set of seed hashtags, for instance: \#dpcm (decree of the president of the council of ministers), \#legge (law) and \#leggedibilancio (budget law), we retrieved a sample of tweets and then added the new hashtags contained in this sample to extend the list of seed hashtags and retrieve new tweets. \hl{We called this dataset Policycorpus}. We removed duplicates, retweets and tweets containing only hashtags and urls. \hl{In total we obtain} a set of 1264 tweets (1000 for training and 264 for testing). The amount of hate labels in the Policycorpus is 11\% (1124 normal and 140 hate tweets). It is strongly unbalanced, like in the it-HS corpus (17\% of hate tweets), because it reflects the raw distribution of hate tweets in Twitter. The HaSpeeDe-tw corpus (32\% of hate tweets) instead has a distribution that oversample hate tweets.  
At the end of the sampling process, the list of seeds included about 60 hashtags referring to
\begin{itemize}
    \item Laws, \hl{such as} \#decretorilancio (\#relaunchdecree), \#leggelettorale (\#electorallaw), \#decretosicurezza (\#securitydecree)
    \item Politicians and policy makers, \hl{such as} \#Salvini, \#decretoSalvini (\#Salvinidecree), \#Renzi, \#Meloni, \#DraghiPremier
    \item Political parties, \hl{such as} \#lega (\#league), \#pd (\#Democratic Party)
    \item Political tv shows, \hl{such as} \#ottoemezzo, \#nonelarena, \#noneladurso, \#Piazzapulita
    \item Topics of the public debate, \hl{such as} \#COVID, \#precari (\#precariousworkers), \#sicurezza (\#security), \#giustizia (\#justice), \#ItalExit
    \item Hyper-partisan slogans, \hl{such as} \#vergognaConte (\#\hl{shameonConte}),  \#contedimettiti (\#ConteResign) or \#noicontrosalvini (\#WeareagainstSalvini)
\end{itemize}  

This is the first corpus in Italian annotated with hate speech against policy makers. We plan to make this resource available under request\footnote{The url is removed during the review process}.

To produce gold standard labels, we asked two Italian experts of communication, to manually label the tweets in the Policycorpus, distinguishing between hate and normal tweets according to the following guidelines:
By definition, hate speech is \textit{any expression that is abusive, insulting, intimidating, harassing, and/or incites to violence, hatred, or discrimination. It is directed against people on the basis of their race, ethnic origin, religion, gender, age, physical condition, disability, sexual orientation, political conviction, and so forth.}
Translated Examples:

1) ``\textit{a clear \#NO to \#Netherlands that we would like users of the \#MES economic resources but in exchange for Italy's renunciation of its budgetary autonomy. To Netherlands we say: thank you and goodbye, WE ARE NOT INTERESTED !!}''
is normal because it does not contain hate, insults, intimidation, violence or discrimination.

2) ``\textit{... There is a weekly catwalk of the \#jackal \#no \#notAtAll! Listening to a Po \#clown after a true PATRIOT a doctor from \#Bergamo cannot be held, seen or heard. Giletti should stop inviting certain SLACKERS FROM THE PO VALLEY! \#COVID-19 \#NonelArena}'' 
contains hate speech, \hl{including} insults like \#clown and \#jackal.

3) ``\textit{I have my say ... \#Draghi is a great economist but we don't need a \#Monti-style economist ... We don't need another technical \#government to obey the banking lobby! We need a political leader! We need a \#ItalExit! We need the \#Lira! \#No to \#DraghiPremier}''
is a normal case, despite \hl{the} strong negative sentiment. It might be controversial for the presence of the term \textit{lobby}, often used in abusive contexts, but in this case, it is not directed against people on the basis of their race, ethnic origin, religion, gender, age, physical condition, disability, sexual orientation or political conviction.

The Inter-Annotator Agreement is \textit{k}=0.53. Although the score is not high, it is in line with the score reported in the literature for hate speech against immigrants (\textit{k}=0.54) \cite{poletto2017hate} and indicates that the detection of hate speech is a hard task for humans. All the examples of disagreement were discussed and an agreement was reached between the annotators. The cases of disagreements occurred more often when the sentiment of the tweet was negative, \hl{this was} mainly due to:

The use of vulgar expressions not explicitly directed against specific people but generically against political choices.

The negative interpretation of hyper-partisan hashtags, such as \#contedimettiti (\#ConteResign) or \#noicontrosalvini (\#WeareagainstSalvini), in tweets without explicit insults or abusive language.

The substitution of explicit insults with derogatory words, such as the word ``circus'' instead of ``clowns''. 
In the next section, we report and discuss the results of the experiments.

\section{Experiments and Discussion}\label{exp}
Our goal is to create models of hate speech \hl{that} automatically predict hateful tweets against policy makers in the Policycorpus. First, we describe the features extracted from text, then we perform in-domain and cross-domain classification, and finally, we conduct feature analysis and \hl{visualize} the hashtag networks. As discussed in Section \ref{prev}, we aim to develop explainable Artificial Intelligence (AI) models, hence we also exploited ML algorithms based on lexical resource (Lex), such as SVM, Adaboost and Random Forests, in addition to more advanced techniques, for instance, neural networks based on the AlBERTo pretrained transformer model.
We ran two different experiments: 
\begin{itemize}
    \item In experiment one, we tried to answer to RQ2, using different algorithms to train models on the existing corpora. We then perform a cross-domain classification, evaluating the predictions trained on HaSpeeDe-tw and it-HS to the Policycorpus test set (Section \ref{exp1});
    \item In experiment two, we tried to answer to RQ1, with a feature analysis to understand which features are best predictors of hate speech in the policy making domain with respect to anti-immigration domain (Section \ref{exp2});
\end{itemize}
\hl{Finally}, to answer RQ3, we visualized the networks of hashtags in order to understand the relationships between topics used in normal and hateful tweets (Section \ref{exp4}).
Before all, we described the features extracted from text.

\subsection{Feature Extraction}
\hl{Building upon} the previous work presented in the literature, we adopted linguistic resources for the extraction of features to use with ML algorithms. In particular, we used:
\begin{itemize}
    \item LIWC \cite{tausczik.pennebaker10}, a linguistic resource available in many languages, including Italian \cite{alparone.al04}, that maps words to 68 psycholinguistics dimensions, such as linguistic dimensions (i.e. pronouns, articles, tense), psychological processes (i.e cognitive mechanisms, sensations, certainty, causation) human processes (i.e. sex, social life, family), personal concerns (i.e. leisure, money, religion, death) and spoken categories (i.e. assent, nonfluencies)
    \item NRC \cite{mohammad2013nrc}, a linguistic resource that maps words to 10 emotion and polarity features, for instance positive words, negative words, anger, anticipation, fear, sadness, joy, surprise, trust and disgust.
    \item Other 22 language-independent stylometric features \cite{celli15lato}, including positive/negative emoticons/emojis, ratio of punctuation, question and expression marks, numbers, operators, links, hashtags, mentions or emails addresses, parentheses, lowercase/uppercase and ratio of repeated bigrams.
\end{itemize}
These dictionaries extract a matrix of 100 features, less sparse than bag-of-words. In addition to this, we used a transformer model trained on Italian tweets: AlBERTo, that extracts a dense matrix of more than 700 embedding features \cite{polignano2019alberto}. 

\subsection{In-Domain and Cross-Domain Classification}\label{exp1}
Hate speech labels are naturally unbalanced, as normal tweets are - fortunately - the large majority, especially in the Policycorpus and it-HS corpus. As this is a natural condition, we chose to keep the labels unbalanced and measure the performances with two metrics: ROC AUC curve, which is insensitive to class imbalance, and weighted-average F-measure that takes into account the difference of performance for the two classes. In this experiment, we trained and tested various algorithms, we used a training-test split as evaluation settings, which is 88\%-12\% in the it-HS corpus, 75\%-25\% in HaSpeeDe-tw2018 and 80\%-20\% in the Policycorpus.

\begin{table}[ht]
\centering \small
\begin{tabular}{lll}
\hline
\textbf{dataset-algorithm} & \textbf{ROC} & \textbf{\textit{F1} }  \\
\hline
\hline
it-HS baseline & 0.5 & 0.7 \\
it-HS Lex+random forest & 0.69 & 0.78 \\
it-HS Lex+Support Vector Machines & 0.51 & 0.76 \\
it-HS Lex+adaboost bayesian net & 0.67 & \textbf{0.79} \\
it-HS AlBERTo+Neural Networks &  \textbf{0.85} & 0.76 \\ 
\hline
\hline
HaSpeeDe-tw baseline & 0.5 & 0.4 \\
HaSpeeDe-tw Lex+random forest & 0.73 & 0.71 \\
HaSpeeDe-tw Lex+Support Vector Machines & 0.63 & 0.72 \\ 
HaSpeeDe-tw Lex+adaboost bayes net & 0.73 & 0.72 \\
HaSpeeDe-tw AlBERTo+Neural Networks & \textbf{0.88} & \textbf{0.82} \\ 
\hline
PC baseline & 0.5 & 0.8 \\
PC Lex+random forest & 0.69 & 0.87 \\
PC Lex+Support Vector Machines & 0.52 & 0.88 \\ 
PC Lex+adaboost decision tree & 0.57 & 0.88 \\
PC AlBERTo+Neural Networks & \textbf{0.83} & \textbf{0.89} \\ 
\hline
\end{tabular}
\caption{\small{Results of the classification of Hate Speech in Italian on the Italian HS corpus (it-HS), HaSpeeDe-tw2018 (HaSpeeDe-tw) and Policycorpus (PC) with different algorithms, lexical features (Lex) and transformer embeddings (AlBERTo). As evaluation metrics we used ROC AUC and weighted average F1 measure.} \label{reslab}}
\end{table}

\begin{table}[ht]
\centering \small
\begin{tabular}{llll}
\hline
\textbf{corpus and label} &  \textbf{precision} &  \textbf{recall} & \textbf{f1-score} \\
\hline
\hline
\hl{PC} normal & 0.92 & 0.98 & 0.95 \\
\hl{PC} hate & 0.76 & 0.41 & 0.54 \\
\hline
it-HS normal & 0.71 & 0.92 & 0.81 \\
it-HS hate & 0.86 & 0.56 & 0.68 \\
\hline
HaSpeeDe-tw18 & 0.86 & 0.89 & 0.88 \\
HaSpeeDe-tw18 & 0.75 & 0.71 & 0.73 \\
\hline
\end{tabular}
\caption{\small{Per-class results of the classification on each corpus with the best algorithm (AlBERTo + Neural Networks).} \label{perclass}}
\end{table}
A closer look \hl{at} the per-class performance obtained with the best algorithm (AlBERTo + neural networks), reveals that in general the algorithm has a higher performance in the detection of normal tweets and lower \hl{performance in} the recognition of hate tweets, which have a poor recall. The fact that recall is higher in the HaSpeeDe-tw corpus than in the Policycorpus suggests that balancing the number of hate examples with the normal ones has a positive effect on recall. Precision is similar in these two datasets (0.75): the it-HS corpus has a higher precision on the hate class, but the recall follows the same pattern of the other two corpora. We \hl{present} these results in Table \ref{perclass}.

\hl{In attempt to address} RQ2, we used the models trained on the HaSpeeDe-tw and it-HS corpora in the previous experiment to automatically produce predictions on the Policycorpus test set, thus performing a cross-domain backtest. Given the differences between domains we expect poor results in the next experiment, \hl{the results of which are presented} in Table \ref{back}. 
\begin{table}[h]
\centering \small
\begin{tabular}{lll}
\hline
\textbf{dataset-algorithm} & \textbf{ROC} & \textbf{\textit{F1} }  \\
\hline
\hline
Pc-test baseline & 0.5 & 0.8 \\
\hline
HaSpeeDe-tw-train to PC-test Lex+random forest & \textbf{0.71} & \textbf{0.77} \\
HaSpeeDe-tw-train to PC-test Lex+SVM & 0.56 & 0.73 \\
HaSpeeDe-tw-train to PC-test Lex+adaboost bayesian net & 0.71 & 0.76 \\
HaSpeeDe-tw-train to PC-test AlBERTo+Neural Networks & 0.61 & 0.72 \\
\hline
it-HS-train to PC-test Lex+random forest & \textbf{0.81} & 0.88 \\
it-HS-train to PC-test Lex+Support Vector Machines & 0.5 & 0.8 \\
it-HS-train to PC-test Lex+adaboost bayesian net & 0.66 & \textbf{0.89} \\
it-HS-train to PC-test AlBERTo+Neural Networks  & 0.53 & 0.82 \\
\hline
\end{tabular}
\caption{\small{Results of the cross-domain classification of Hate Speech in Italian on the Policycorpus-test (PC-test) with the models trained on the HaSpeeDe-tw2018 and Italian HS corpora.} \label{back}}
\end{table}
As expected, the results of cross-domain classification show that the domain shift had a huge impact on the performance of the classifiers, \hl{particularly} from HaSpeeDe-tw to Policycorpus, where the results measured with weighted-average F1 are below the majority baseline, suggesting that the features are so different that the model cannot use them in the correct way. Surprisingly, the models trained on the it-HS corpus produced good results, but only the ones trained with ML algorithms, \hl{particularly} random forests and adaboost, that are more capable of using weak features. AlBERTo and Neural Networks in this case performed only slightly better than the majority baseline. We believe that the large training size of it-HS corpus had a positive effect for the cross-domain adaptation.

\subsection{Feature Analysis}\label{exp2}
The cross-domain classification highlighted the difference in the features between the corpora. To measure this difference, and answer RQ1, we computed the Pearson correlation between the lexical features and the hate speech scores. In Table \ref{corr} we \hl{present} the best lexical features correlated to hate speech in each dataset. Positive correlation indicates the best features to classify hate messages and negative correlations indicate the best feature to classify normal messages. All these features were used in the classification experiments. 
\begin{table}[h]
\centering \small
\begin{tabular}{ll}
\hline
\textbf{Lex feature} & \textbf{best corr. to HS} \\
\hline
\hline
 \textit{HaSpeeDe-tw} &  \\
 lower case (style) & +0.25* \\
 puntuation to word ratio (style) & +0.18 \\
 expression marks (style) & +0.17* \\ 
 uppercase char ratio (style) & -0.15**\\
 numbers (style) & -0.21* \\ 
 non-ending punctuation (style) & -0.3* \\ 
\hline
 \textit{it-HS} &  \\
 expression marks (style) & +0.11** \\
 certainty (LIWC) & +0.1* \\ 
 anger (LIWC) & +0.1** \\ 
 prepositions (LIWC) & -0.05* \\ 
 articles (LIWC) & -0.05* \\
 ending punctuation (style) & -0.05**\\
\hline
 \textit{Policycorpus} &  \\
 swears (LIWC) & +0.15* \\
 initial uppercase (style) & +0.15** \\
 sexual (LIWC) & +0.13** \\ 
 upper case (style) & +0.12** \\ 
 positive emotions (LIWC) & -0.05**\\
 numbers (style) & -0.06*\\
 lower case (style) & -0.07*\\
\hline
\end{tabular}
\caption{\small{Results of the correlation ranking between different lexical features and hate speech. **=P-value lower than 0.01, *=P-value lower than 0.05} \label{corr}}
\end{table}

\begin{figure*}[ht]
\centering \includegraphics[width=\textwidth]{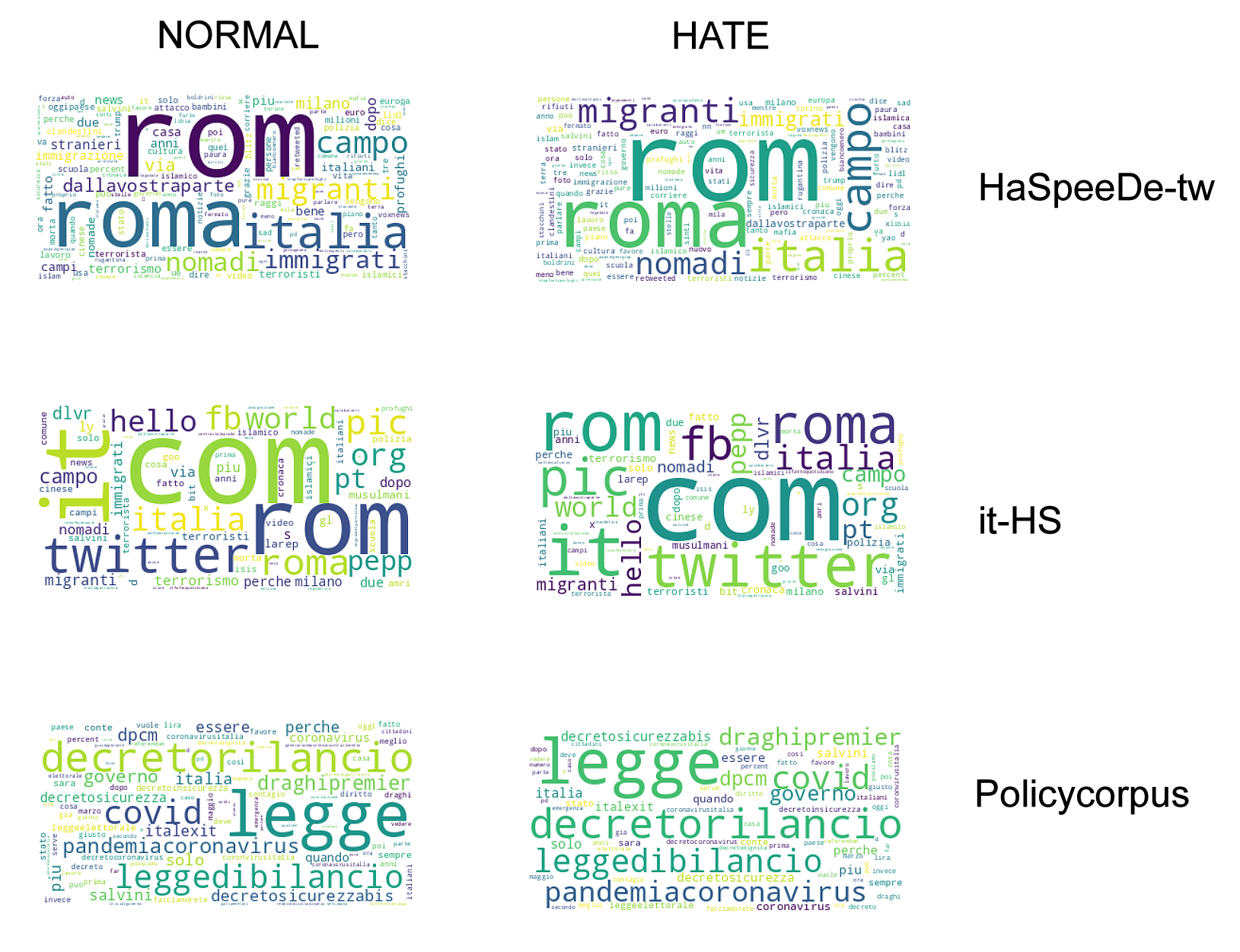}
\caption{\small{Visualization \hl{of} the average activations of each token in the attention vectors, associated to hate and normal labels, for each corpus}}\label{fig0}
\end{figure*}

The analysis revealed that Stylometric features, such as the ratio of lowercase and uppercase characters, have a strong predictive power in the HaSpeeDe-tw2018 dataset, but not in the it-HS corpus, where there is more variety. LIWC features, such as sexual, anger and swear word ratios, are among the best predictors of hate speech against politicians. This experiment clearly shows that the most useful features for the detection of hate speech in the domain of anti-immigration are punctuation (the more there is punctuation, the more a message is non-hateful) and expression marks (the more exclamations, the more a message is likely to be hateful). In Policycorpus there are sexual and swear words as markers of hateful messages and lower case, numbers and positive emotions as markers of non-hateful messages. It is interesting to note that lower case letters are correlated to hate speech in the anti-immigration domain, while in the anti-policy domain they are correlated to non-hateful m\hl{e}ssages. The similarity between the best features in it-HS and Policycorpus explained the good result obtained in the cross-domain classification with ML algorithms.
We also exploited the attention vectors of AlBERTo to try to explain the poor performance in the cross-domain classification. Using the average activations of each token in the attention vectors, we computed the strongest predictors in the model. The results, represented as wordclouds in Figure \ref{fig0}, show the most frequent tokens activated to detect hate and normal labels for each corpus. The clear difference from the tokens used in the anti-immigrant and anti-policy domains is a clue of the poor performance in cross-domain classification.

\subsection{Hashtags Network Analysis}\label{exp4}
\begin{figure*}[h]
\centering \includegraphics[width=\textwidth]{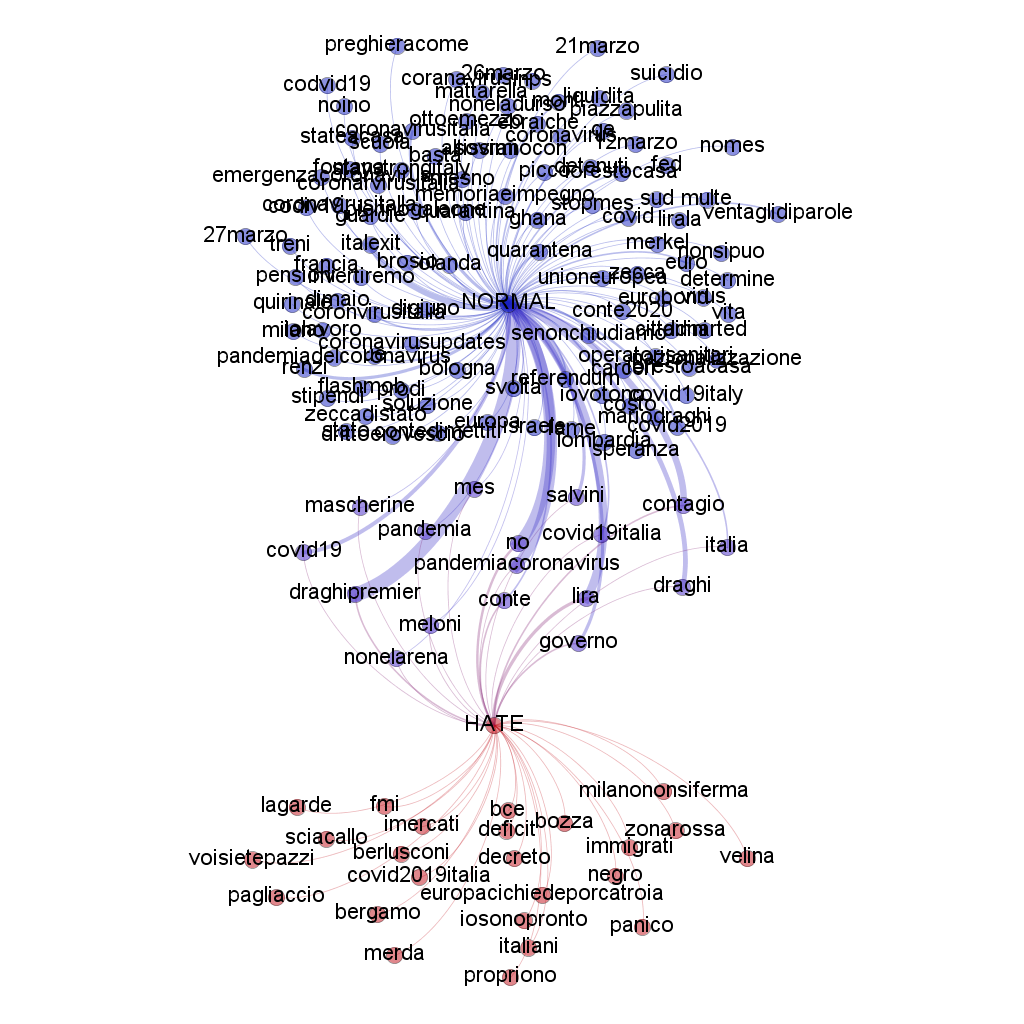}
\caption{\small{Detailed visualization of the hashtag network of a small portion of the Policycorpus with Yifan Hu trees. The blue cloud (above) contains hashtags of normal messages, the red cloud (below) contains the hashtags from hate tweets and the smaller cloud between the two contains the hashtags in both messages.
}}\label{fig2}
\end{figure*}
To \hl{address} RQ3, we treated the `normal' and `hate' classes as nodes in a network that we plotted with Yifan Hu trees \cite{hu2015visualizing}. In this way we were able to visualize the network of hashtags connected to the `normal' or `hate' nodes in the Policycorpus. In other words, we were able to visualize hashtags appearing only in hate speech context, hashtags appearing only in normal contexts, and hashtags appearing in both networks. The results, depicted in Figure \ref{fig2}, show a pattern with a blue cloud (above), that represent\hl{s} the network of hashtags in normal tweets and a red cloud (below), that represents the hashtags in hate messages. Between the two, there is a smaller cloud of hashtags used in both contexts. A closer look to these hashtags reveals the topics of the public debate that are more controversial. 

These topics include politicians (\#Salvini, \#Meloni, \#Conte, \#Draghi), economic issues (\#lira, \#MES), keywords related to the pandemic (\#covid, \#pandemia, \#mascherine) and to political tv shows (\#nonelarena). 


\section{Conclusion and Future Work}\label{conc}
In this paper, we presented a new resource for the analysis of hate speech against policy makers on Twitter. The dataset, named Policycorpus, is the first \hl{of this type} in Italian, an under-resourced language. We confirmed that the annotation of hate speech is \hl{difficult}, and detailed the cases of disagreements between annotators. Using this resource, we demonstrated that:
\begin{itemize}
    \item Deep Learning algorithms and transformer-based models achieve state-of-the-art performances in both domains.
    \item Machine Learning algorithms are suitable for cross-domain classification from hate speech against immigrants to hate speech against policy makers.
    \item Hate speech against immigrants can be detected \hl{by} looking at the style of the written text (i.e. punctuation \hl{and exclamation}), while hate speech towards policy makers is based more on the vocabulary and psycholinguistic aspects (i.e. swear words).
\end{itemize}
We also visualized the spread of hate speech in Twitter against policy makers \cite{hagen2019open} and identified clusters of tweets that appear only in hate tweets and in both normal and hate tweets. We suggest that this method can be exploited to track which topics convey hatred towards policy-makers. Combining hate speech detection algorithms and visualizations, one can build a dashboard for monitoring hate speech on Twitter. The \hl{final} aspect that we want to highlight is that the amount of data available, and its balance between classes, can help \hl{to improve} the performance of the classifiers.
In the future we plan to run experiments \hl{on} domain-adaptation and collect more data for hate speech detection against policy makers.

\section*{Conflict of Interest}
Three of authors of this paper declare to be employed by Maggioli S.p.A., a private Company with a financial interest in the field of Public Administration. They also declare their involvement in the PolicyCLOUD project, which has received funding from the European Union’s Horizon 2020 research and innovation programme (Grant Agreement no 870675), and has financial interests in the subject and materials discussed in this manuscript.

\section*{Data Availability}
The authors plan to make Policycorpus, the dataset presented in this manuscript, available upon request, under the conditions set by the PolicyCLOUD project. For more information, please get in touch with fabio.celli@maggioli.it.

\bibliographystyle{plain}
\bibliography{my}



\end{document}